# Exploiting a Fleet of UAVs for Monitoring and Data Acquisition of a Distributed Sensor Network

S. MahmoudZadeh, A. Yazdani, A. Elmi, A. Abbasi, P. Ghanooni

*Abstract*— This study proposes an efficient data collection strategy exploiting a team of Unmanned Aerial Vehicles (UAVs) to monitor and collect the data of a large distributed sensor network usually used for environmental monitoring, meteorology, agriculture, and renewable energy applications. The study develops a collaborative mission planning system that enables a team of UAVs to conduct and complete the mission of sensors' data collection collaboratively while considering existing constrains of the UAV payload and battery capacity. The proposed mission planner system employs the Differential Evolution (DE) optimization algorithm enabling UAVs to maximize the number of visited sensor nodes given the priority of the sensors and avoiding redundant collection of sensors' data. The proposed mission planner is evaluated through extensive simulation and comparative analysis. The simulation results confirm the effectiveness and fidelity of the proposed mission planner to be used for the distributed sensor network monitoring and data collection.

## 1. Introduction

The salient features of mobility, maneuverability, reconfigurability, and autonomy make UAVs as an enabling technology to be used in a wide range of applications such as field inspection and monitoring, remote sensing, surveillance, interrogation, and data collection [1-3]. On the other hand, development of infrastructures such as sensor networks (SNs) have been growing quickly in the era of Industry 4.0. This infrastructure is very useful in capturing and collecting important data which can be used for environmental changes analysis, meteorology prediction, and communication purposes. However, there still exist some shortages with the physical design and current technologies of communication protocols in these systems which cannot allow efficient and persistent broadcasting of the network data to the central station. For example, the sensor nodes in a network located in a remote area are usually supplied with the battery that is difficult to be recharged regularly. This limited capacity of the battery charge has direct impact on persistent transmission of data to the sink node and then central station demanded a high amount of energy per transmission [2]. On top of that, constraints associated with sensors' buffer do not allow sensor to maintain the data packet for a long time and the buffer overflow and losing data are the negative consequences of this situation [3]. As a result, UAVs as a portable router technology can be employed to enhance persistent transmission of sensors' data and avoid losing the data packets.

There are several documented research studies in the state-of-the -art employed UAVs to leverage the performance of SNs.

In [4], a fixed-wing UAV is employed as a mobile data collector to serve the scattered ground terminals using a cyclical approach. However, this research considers terminals to be distributed uniformly on a straight line with the same distance from each other, which is not applicable in real world scenarios. Furthermore, a fixed-wing UAV is not able to hover; hence, the study assumes the UAV flies repeatedly in horizontal plane with a fixed altitude, which restricts the vehicle's maneuverability to read sensors' data in the predefined area. In [5], the link between SNs and UAV is modelled as a general fading channel aiming at reducing packet loss and minimizing SNs' energy consumption by jointly optimizing SNs' sleep/wake-up mechanism and UAV's trajectory. In this study, however, the sensors residual energy and data collection frequency is not addressed. In [6], a rotary-wing UAV is employed to collect data in a SN. The proposed UVA used in this work is capable of hovering over a fixed location, enabling a better communication channel to the sensors. This effectively addresses the power consumption minimization problem of the vehicle-sensor communication subjected to the targeting rate constraint. In [7], a UAV-enabled age-optimal data collection approach is proposed. This work considers the data upload timestamp and the time elapsed since the UAV gets the data. In this study, the problem of collecting data of a SN using a single UAV is modelled as a shortest Hamiltonian path and is solved by a dynamic programming approach. In [8], a UAV route planning approach based on the mixed-integer programming while considering the sensors sleep-wake-up schedule is developed to mitigate the energy consumption of a SN. . The simulation results of this study confirm the effectiveness of the proposed route planning approach although an experimental trial is yet unexplored. In [9], a multi-UAVs strategy is proposed to address the problem of optimum data collection from distributed sensors. In this study, the UAVs' propulsion power consumption as well as the ground sensors' power consumption are considered as the primary performance indices. This study proposes an iterative approach to leverage successive convex approximation of the multiple UAVs route planning. In [10], a multi-agent reinforcement learning

Somaiyeh MahmoudZadeh is with the School of IT, Deakin University, Geelong, VIC 3220, Australia (e-mail: s.mahmoudzadeh@deakin.edu.au)

Amirmehdi Yazdani is with the College of Science, Health, Engineering and Education, Murdoch University, Perth, WA 6150, Australia (e-mail: amirmehdi.yazdani@murdoch.edu.au)

Atabak Elmi is with the School of IT, Deakin University, Geelong, VIC 3220, Australia (e-mail: atabak.elmi@deakin.edu.au)

Amin Abbasi is with the Department of Electrical Engineering,Azad University of Khoemeinishar, Esfahan, Iran (e-mail: aminabbasi.res@gmail.com)

Pooria Ghanooni, Department of Electrical Engineering, Azad University of Mashhad, Mashhad, Iran (e-mail: pooria.ghanooni@gmail.com)

framework for dynamic resource allocation of UVAs in a SN is developed. In this work, the local observation and learning are used for each agent to explore the best strategy of communication and data collection. Simulation results of this work demonstrate that a tradeoff between the information exchange overhead and the system performance for multi-UAV networks can be achieved.

In [11], a UAV-aided data gathering approach in SNs is proposed considering a three-stage sensor transmission policy, where the main focus is on the optimization of the packet transmission energy. This study dynamic programming algorithm to attain the best transmission policy, while the UAV follows a preplanned trace obtained by recursive random search algorithm. Although the proposed approach is energy efficient in terms of sensor data transmission, the UAV is restricted to take the preplanned optimum route, which might not be valid or optimum in one step forward especially in the cases of scalable networks.

Finally, in [12], UAVs-based data collection in a dense wireless SN is proposed. In this study, a projection-based compressed data gathering is utilized, and the UAV-based data collection problem is formulated as a joint optimization problem comprising sensor nodes clustering, cluster head selection, and UAV trajectory planning. The simulation results show that the proposed approach significantly reduces the number of transmissions in the network and consequently leveraging the energy saving and the network lifetime.

Considering all mentioned works, this paper aims to further investigate the limitations of efficient and sustainable SN's data transmission and to propose a collaborative UVA-based solution to address the issues. The sensors have limited coverage and energy capacity that can be faded away from the network when their battery dies. Hence, the network topology can change by time as some sensors can get discarded due to loss of energy, and some sensors can be added to the network. On the other hand, data collection in a network of heterogeneous sensors, where the sensors have diverse collection/transmission priorities and operational frequencies, and are scattered at distant positions where no fixed gateway coverage or internet is available, is quite challenging. This requires appropriate prioritizing the order of sensors to be visited within a limited time frame.

Although a UAV can flight sufficiently close to the sensor nodes and can use of the line-of-sight (LoS) communication links for data collection in a SN, however, the UAV deployment and mission planning optimization for persistent and efficient data collection in the SN are new challenges that should be investigated.

In order to address these challenges, this research proposes the application of multiple UAVs for data collection in a SN. For obtaining maximum data collection in a single mission, this paper develops a collaborative mission planning system that enables a team of UAVs to conduct and complete the mission of sensors' data collection collaboratively while considering existing constrains of the UAV payload and battery capacity. The proposed mission planner system employs the DE optimization algorithm enabling UAVs to maximize the number of visited sensor nodes given the priority of the sensors and avoiding redundant collection of sensors' data. The outcome of this paper can effectively contribute to minimize the transmission energy consumption and the sensory data packet loss that results in prolonging the lifetime of the SN.

The rest of this paper is organized as follows. In Section 2, the multiple UAVs routing and data collection is mathematically formulated. Section 3 discusses the application of DE algorithm on the UAVs collaborative mission planning. Discussion on the simulation results and a comparative assessment of multi UAVs collaborative and non-collaborative mission planning is provided in section 4. Section 5 concludes the paper.

**2. Problem Formulation**

A heterogeneous network of sensors with non-uniform scattering pattern is considered for this study. This means that the sensors are not in a straight line and transmission frequency and residual energy consumption of sensors are not identical. The mission of data collection is conducted by collaborative work of 4 UAVs in this study. A sample of such a network is depicted by Fig.2. The UAVs receive the topology of SN, position of sensors, and adjacency information of the network a priori. More specifically, the mission of the UAVs is defined as:

*Depurating from the base station, collecting the data of prompted sensors, and then arriving to the docking station within the given mission time and before reaching the battery state of the charge to a pre-defined threshold.*

The main objective of the mission is to maximize the data collection in a single mission that corresponds to maximize the number of sensor nodes that should be visited. To this end, the SN can stay idle unit they receive the "*Hello Message*" from UAVs. Then, they can communicate and transmit the data to the UAVs by using the LoS communication link. In this regard, the UAVs' mission planner should consider the priority of the sensors according to their frequency of transmission, residual battery, and priority of the data to be collected, and then generate optimal routes to guide the UVAs to maximize the number of sensor node visited. Finally, the mission planner should transverse the UAVs to the docking station before the UVAs' battery charge becomes below the threshold.

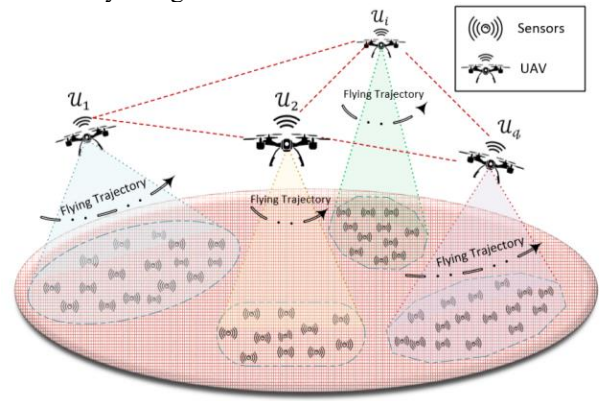

Fig.2. Illustration of multi-UAVs collaborative sensor data collection.

Existence of prior information about the environment, and geographical position of sensors, has a significant impact on UUV's efficient and robust mission planning. A partial map of the Mount Dandenong area with the coordinates of ⟨37° 49' 40" S, 145° 21' 09" E ⟩ to ⟨37°49′36″ S, 145°21′17″ E⟩ is provided for the UAVs operation, as presented in Fig.3.

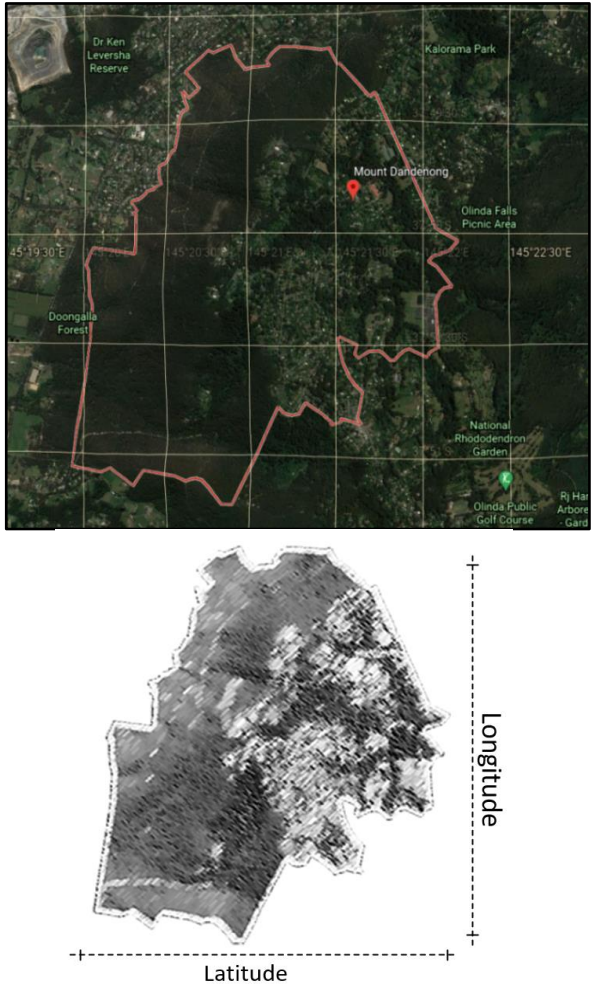

Fig. 3. A snapshot of selected map area in Mount Dandenong area with the latitude and longitude of ⟨37° 49' 40" S, 145° 21' 09" E⟩ to ⟨37°49′36″ S, 145°21′17″ E⟩.

The algorithm uses the geographical map of area with the dimension of 3.4-by-3.4 $km^2$ encapsulating a distributed SN of 128 sensor nodes. The network topology of the SN is also configured and deformed in the operation area. To fulfil the operational requirements, the UAVs are equipped with a set of navigation sensors, RF capability, and camera. To conduct the mission successfully, the sensors' distribution, coverage, residual energy, timestamp, data read priority and frequency of transmission as well as the number and scalability of the sensors to be deployed should be considered. The UAVs can localize the planted sensors, map the network topology, and update the map during deployment while detecting newly integrated or died off sensors.

For collecting the sensory data, an optimal mission planning strategy for the fleet of UAVs will be conducted to collect maximum sensors' data collaboratively, considering overwhelming volume, various priority, available resource, and different read cycle of the planted sensors.

Assuming there are a homogeneous fleet of $q$ identical UAVs $\mathcal{U} = \langle \mathcal{U}_1, ..., \mathcal{U}_q \rangle$, each vehicle will be required to plan the efficient path to collect maximum amount of sensors' data and then returning to the docking station. The UAVs collect the data from each bunch of locally distributed sensors by selecting an appropriate collection points (CPs) at which the UAV hovers to collect sensed data based on the affinity propagation method [13], and schedules which bunch of sensors to upload at each CP. Let $G = (\mathcal{Cp}, \mathcal{E})$ be a graph, representing senor network, in which the $\mathcal{Cp}$ corresponds to the collection points (nodes) that is presented by $\mathcal{Cp} = \{S_{xy}^1, ..., S_{xy}^j, S_{xy}^l ..., S_{xy}^n\}$, where each $S_{xy}^j$ is an arbitrary sensor in the geographical frame $x - y$ for the UAV to hover and collect the detailed data; $\mathcal{E}^{S^{jl}} = \{\{S_{xy}^j, S_{xy}^l\} | S_{xy}^j, S_{xy}^l \in \mathcal{Cp}, S_{xy}^j \neq S_{xy}^l\}$ is a set of edges. Due to energy restrictions of the vehicles and sensors, and extensive number of sensors distributed in a large operation field, completing all tasks in one mission is not feasible for a limited number of vehicles. Therefore, an impact factor of $\rho$ has been assigned to each sensor to prioritize the order of collection which should be completed and govern the vehicles toward the destination. In this context, the UAV mission planner simultaneously tends to determine the optimum order of collection points to be visited mathematically described as follows:

$$\mathcal{Cp} = \{S_{xy}^1, ..., S_{xy}^j, ..., S_{xy}^n\};$$
$$\forall S_{xy}^j \in \mathcal{Cp}, \exists \rho^j \sim \mathbb{U}(1,100)$$
$$Bound_{xy}^{max} \sim \mathbb{N}(0, \sigma^2)$$
$$|S_{xy}^j| \leq Bound_{xy}^{max}$$
(1)

Here, each sensor is associated with a priority value of $\rho^j$. The points should be distributed in a respective search region of the map. Hence, the region is constrained with a predefined bound of $Bound_{xy}^{max}$ in the x-y Cartesian coordinates, which is equal to size of the map.

Here, we represent it as a CP matrix, where each point has weight and completion time.

$$\mathcal{Cp}_{xy} = \begin{bmatrix} S_{x_1,y_1}^1 & S_{x_1,y_2}^2 & \cdots & S_{x_1,y_n}^n \\ S_{x_2,y_1}^2 & S_{x_2,y_2}^2 & \cdots & S_{x_2,y_n}^n \\ \vdots & \vdots & \vdots & \vdots \\ S_{x_n,y_1}^n & S_{x_n,y_1}^n & \cdots & S_{x_n,y_n}^n \end{bmatrix},$$
(2)

$$\{S_{x,y}^j \in \mathcal{Cp}_{xy} | j = 1, ..., n\}$$

where $n$ is the number of distributed sensors, $j$ is the index of each sensor. Travel interval between node $S_{xy}^j$ and $S_{xy}^l$ is introduced by $\mathcal{D}^{S^{jl}} = \sqrt{(S_x^l - S_x^j)^2 + (S_y^l - S_y^j)^2}$ and the corresponding travel time is indicated by $T^{S^{jl}}$. Accordingly, we will have a distance matrix to represent the pairwise distance between the sensors that is presented in Fig.4.

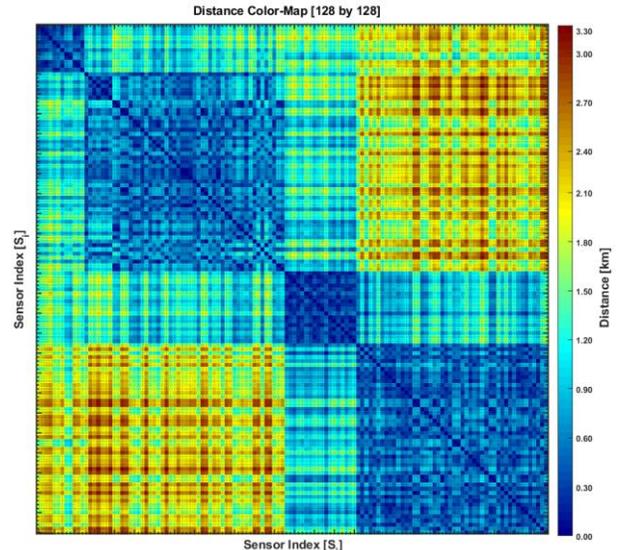

Fig.4. Pairwise distance between 128 sensors, where the color bar represents the distance starting from 0.0 (*km*) to 3.4 (*km*) performed by different colors (From blue for minimum to red for maximum distance between a pair of sensors).

Let us assume that each UAV $\mathcal{U}_i$ has a maximum energy capacity of $\mathcal{T}_{\mathcal{U}_i}^{\max}$, and $\mathcal{T}_{\mathcal{U}_i}$ is the energy consumed by $\mathcal{U}_i$ per time unit. Time that $i^{th}$ UAV arrives at the node $\mathcal{S}_{xy}^j$ is denoted by $t_{\mathcal{U}_i}^{\mathcal{S}_{xy}^j}$ and Ground speed of a UAV from node $\mathcal{S}_{xy}^j$ to $\mathcal{S}_{xy}^l$ is $\mathcal{V}g_{\mathcal{U}_i}^{\mathcal{S}^{jl}}$. Incorporating the network configuration, estimated time for visiting the set of selected sensors should be fitted to the time constraint of the UAVs battery lifetime $\mathcal{T}_{\mathcal{U}_i}^{\max}$. Assuming a binary variable of $\alpha_{\mathcal{U}_i}^{\mathcal{S}^{jl}}$ denoting whether or not the vehicle $\mathcal{U}_i$ travelled from $\mathcal{S}_{xy}^j$ to $\mathcal{S}_{xy}^l$:

$$\alpha_{\mathcal{U}_i}^{\mathcal{S}^{jl}} = \begin{cases} 1 & if\ \mathcal{U}_i\ travels\ from\ \mathcal{S}_{xy}^j\ to\ \mathcal{S}_{xy}^l \\ 0 & Otherwise \end{cases} \quad (3)$$

The UAVs are given a map of the environment prior to the mission. During the mission, each vehicle should exchange the information of its pose, sensors coordinate, and number of tasks completed with other vehicle. In a case that a vehicle aborts its mission for any reason (e.g., ran out of battery), the closest vehicle undertakes the incomplete mission, or the mission is rearranged between several vehicles and a new mission scenario for all vehicles is planned.

The proposed mission planning problem in this study is modeled as, a fixed-time optimal control problem according to battery capacity. The main objective is to maximize the total number of the met sensors for a fleet UAVs, while addressing the order of visit according to the sensors priority value $\rho^j$. The problem also has some strict constraints to be considered. Given that $t_0$ is the time spent for take-off and landing of a vehicle, if $\mathcal{U}_i$ is traveling from $\mathcal{S}_{xy}^j$ to $\mathcal{S}_{xy}^l$, then arrival time $t_{\mathcal{U}_i}^{\mathcal{S}_{xy}^l}$ to $\mathcal{S}_{xy}^l$ must be equal to the sum of flight time between node $\mathcal{S}_{xy}^j$ to $\mathcal{S}_{xy}^l$ and the arrival time $t_{\mathcal{U}_i}^{\mathcal{S}_{xy}^j}$ to node $\mathcal{S}_{xy}^j$ including the time spent for take-off $t_0$. The main constraint is to ensure that the demand assigned to $\mathcal{U}_i$ does not exceed its maximum energy capacity of $\mathcal{T}_{\mathcal{U}_i}^{\max}$. Another constraint is to ensure that all UAVs $\mathcal{U}_i$ arrive to the same rendezvous point, as the final destination, before reaching to the $\mathcal{T}_{\mathcal{U}_i}^{\max}$. This means that the energy consumed by the $i^{th}$ UAV should be less than or equal to the battery limit of the $\mathcal{U}_i$.

$$\mathcal{Q}_{total} = \min \sum_{i=1}^{q} \sum_{j=1}^{n} \sum_{l=1}^{n} \left( \frac{\mathcal{D}_{\mathcal{U}_i}^{\mathcal{S}^{jl}}}{\mathcal{V}g_{\mathcal{U}_i}^{\mathcal{S}^{jl}}(\mathcal{S}^{jl}, \rho^j, \rho^l)} \right) \times \alpha_{\mathcal{U}_i}^{\mathcal{S}^{jl}}$$

$s.t$

$$\begin{cases} \forall \mathcal{S}_{xy}^j, \mathcal{S}_{xy}^l \in \mathcal{C}p \\ \forall\ i \in q \end{cases} ; \left( \alpha_{\mathcal{U}_i}^{\mathcal{S}^{jl}} = 1 \right) \quad (4)$$

$$\mapsto \left( t_{\mathcal{U}_i}^{\mathcal{S}_{xy}^l} = T^{\mathcal{S}^{jl}} + \sum_{k=1}^{j} t_{\mathcal{U}_i}^{\mathcal{S}_{xy}^k} + t_0 \right)$$

$$\sum_{\mathcal{S}^j \in \mathcal{C}p} \sum_{\mathcal{S}^l \in \mathcal{C}p} \alpha_{\mathcal{U}_i}^{\mathcal{S}^{jl}} t_{\mathcal{U}_i}^{\mathcal{S}_{xy}^l} \leq \mathcal{T}_{\mathcal{U}_i}^{\max}; i = 1 \dots q$$

where $\alpha_{\mathcal{U}_i}^{\mathcal{S}^{jl}}$ is the binary decision variable, $T^{\mathcal{S}^{jl}}$ is the travel time from node $\mathcal{S}_{xy}^j$ to $\mathcal{S}_{xy}^l$ traversing distance of is $\mathcal{D}^{\mathcal{S}^{jl}}$. $\mathcal{Q}_{total}$ is the objective function of the problem.

In this study, a precedence-based k-means clustering method is employed for grouping the distributed sensors according to their priority of collection and the distance matrix. The highest priority sensors are grouped in smaller clusters to ensure there will be sufficient time for the assigned UAV to serve them.

Given a set of $n$ sensors nodes $\mathcal{C}p = \{\mathcal{S}_{xy}^1, \dots, \mathcal{S}_{xy}^j, \mathcal{S}_{xy}^l \dots, \mathcal{S}_{xy}^n\}$, where each sensor is a 2-dimensional real vector indicating its position on the map, the k-means algorithm aims to partition sensors into $k (\leq n)$ groups $g: \{g_1, \dots, g_k\}$ to minimize the within-cluster variance. In this research, the number of SN-groups $(k)$ is set to be equal to the number of UAVs in the fleet $(q)$. Each senor node $\mathcal{S}_{xy}^j \in \mathcal{C}p$ associated with a priority value of $\rho^j$ and periodically reads and reports its measurement. Accordingly, the k-means used Euclidean distance of the sensors $\mathcal{D}^{\mathcal{S}^{jl}}$ and their priority of collection $(\rho^j)$ to group up the sensors as follows:

$$\arg\min_g \left\{ \sum_{i=1}^{k} \frac{1}{2|g_i|} \sum_{\substack{\mathcal{S}^j, \mathcal{S}^l \in g_i \\ \mathcal{S}^j \neq \mathcal{S}^l}}^{n} \left\| \frac{\mathcal{D}^{\mathcal{S}^{jl}}}{\rho^j, \rho^l} \right\|^2 \right\} \quad (5)$$

## 3. Differential Evolution-Based Collaborative Mission Planning of UVAs

Multiple UAVs' collaborative operations can considerably minimize the cost of deployment, launch and recovery, and boost the effectiveness of the missions restricted by a UAV's endurance and battery capacity. This is a complex NP-hard optimization problem with a strict time restriction of which solving is a nontrivial process that brute force algorithms might not be computationally fast enough to tackle the real-time action-reaction requirements of the problem [14]. Currently, no polynomial-time algorithm is known yet to solve such a complex NP-hard problem [15], and meta-heuristics are proposed as viable candidates to deal with the complexity of these problems as suggested by numerous studies in the area [16, 17].

In this research, DE is conducted to produce an operation plan for multiple UAVs collaborative mission planning in a large-scale environment. The DE is a population-based heuristic optimization approach offering three advantages of secure and fast convergence, adopting comparatively fewer control parameters and attaining the global optima in multimodal search spaces [18]. This method is applicable to a wide range of real-valued problems over continuous space.

DE offers similar operators of which the GA tends to use, with the main difference in mutation mechanism that is using the variance of two solution vectors yields a difference vector with an integrated scaling factor to traverse the search space. The DE algorithm uses floating-point real numbers for coding the problem parameters and adopts non-uniform crossover and differential mutation operations, which results in enhanced solution quality and faster convergence [19]. This provides a flexible ground for parameters' penalty-tuning, and shuffling and evolution of the existing population to the desired solution space. Thus, with the clarity gained by hindsight from the applications, the DE algorithm is employed and tuned for the proposed multi UAVs collaborative route planning problem in this paper. In the following, the basic operators of DE and procedural steps of its evolution process are described.

### *i) Solution Encoding*
The proper vector coding is the main factor of the algorithm's success that directly impacts the quality of the outcome and the algorithm's performance. Given the sensor network topology and adjacency guiding information, a solution vector corresponds to permutation and arrangement of the sensors in the area assigned to vehicles $(\mathcal{S}_{xy}^j)$. Respectively, a feasible solution starts with the coordinates of the start station and ends with the position of the docking station, which can be considered as the same point for UAVs departure and arrival. Assuming the finite set of sensors represented by $\mathcal{C}p =$

$\{S_{xy}^1, \ldots, S_{xy}^j, \ldots, S_{xy}^n\}$, a permutation of the included sensors can be demonstrated as:

$$\chi = \begin{pmatrix} S_{xy}^1, \ldots, S_{xy}^n \\ \gamma_{S_{xy}^1}, \ldots, \gamma_{S_{xy}^n} \end{pmatrix} \quad (6)$$

where, $\langle \gamma_{S_{xy}^1}, \ldots, \gamma_{S_{xy}^n} \rangle$ is a certain arrangement of the sensors $\{S_{xy}^1, \ldots, S_{xy}^n\}$. For mission planning purpose, any arbitrary path permutation $\chi$ in considered as a solution vector of $\chi_i$ ($i=1,\ldots, i_{\max}$) where the elements of the solution vector correspond to sensors of $S_{xy}^j$, and $i_{\max}$ is population size of $\chi_i$. A candidate solution vector is defined as:

$$\chi_{i,t} = \begin{pmatrix} \chi_{1,1} & \chi_{2,1} & \chi_{3,1} & \cdots & \chi_{i_{\max},1} \\ \chi_{1,2} & \chi_{2,2} & \chi_{3,2} & \cdots & \chi_{i_{\max},2} \\ \vdots & \vdots & \vdots & \vdots & \vdots \\ \chi_{1,t_{\max}} & \chi_{2,t_{\max}} & \chi_{3,t_{\max}} & \cdots & \chi_{i_{\max},t_{\max}} \end{pmatrix} \quad (7)$$
$i = 1, \ldots, i_{\max}$
$t = 1, \ldots, t_{\max}$

where, $\chi_{i,t}$ is the $i$-individual of the population in the $t^{th}$ the evolution generation and $t_{\max}$ is the maximum number of iterations (evolution generation).

*ii) Mutation*

The mutation is an evolution operator that produces mutation vectors $\mathcal{M}_{i,t}$, based on the current population $\{\chi_{i,t} | i = 1, \ldots i_{\max}\}$ at generation $t$. Indeed, mutation in DE is the main wheel behind its powerful performance that randomly selects three population members of $\chi_{r1,t}, \chi_{r2,t},$ and $\chi_{r3,t}$ from the same generation $t$, and applies a weighted difference vector between two of them to the third one, which is known as *donor*. Having an appropriate *donor* also accelerates the solution convergence. This study uses a randomly determined donor which is one of the triplet of $\chi_{r1,t}, \chi_{r2,t},$ and $\chi_{r3,t}$, as follows:

$$donor = \sum_{i=1}^{3} \left( \frac{\chi_{ri,t} \lambda_i}{\sum_{i=1}^{3} \lambda_i} \right) \quad (8)$$

where $\lambda_i \in [0,1]$ is a uniformly distributed value. This scheme provides a better distribution of the engaged solution vectors. Thus, the mutant solution vector is produced by:

$$\dot{\chi}_{i,t} = \chi_{r3,t} + \mathcal{F}_\mathcal{M}(\chi_{r1,t} - \chi_{r2,t})$$
$r1 \neq r2 \neq r3$
$r1, r2, r3 \in \{i = 1, \ldots i_{\max}\}$ \quad (9)
$\mathcal{F}_\mathcal{M} \in [0, 1+]$

where, $\mathcal{F}_\mathcal{M}$ is a scaling factor that regulates the amplification of the difference vector and assigning a proper value to $\mathcal{F}_\mathcal{M}$ enhances the exploration ability of DE.

*iii) Crossover*

The mutant and parent vectors $\dot{\chi}_{i,t}$ and $\chi_{i,t}$ are then transferred to the crossover operation. The generated offspring $\ddot{\chi}_{i,t}$ from the crossover operation is represented by:

$$\begin{aligned} \chi_{i,t} &= \langle \gamma_{S_{xy}^1}, \ldots, \gamma_{S_{xy}^n} \rangle \\ \dot{\chi}_{i,t} &= \langle \dot{\gamma}_{S_{xy}^1}, \ldots, \dot{\gamma}_{S_{xy}^n} \rangle \\ \ddot{\chi}_{i,t} &= \langle \ddot{\gamma}_{S_{xy}^1}, \ldots, \ddot{\gamma}_{S_{xy}^n} \rangle \end{aligned} \Bigg| \begin{aligned} \forall j' \in 1, \ldots, m; \quad m \in [1, i_{\max}] \\ \Rightarrow \ddot{\chi}_{i,t} = \begin{cases} \alpha_{j'} \dot{\chi}_{i,t} & \alpha_{j'} \leq \mathcal{F}_c \vee j' \\ \beta_{j'} \chi_{i,t} & \beta_{j'} \leq \mathcal{F}_c \wedge j' \end{cases} \end{aligned} \quad (10)$$

where, $\langle \gamma_{S_{xy}^1}, \ldots, \gamma_{S_{xy}^n} \rangle$ is a certain arrangement of the sensors $\{S_{xy}^1, \ldots, S_{xy}^n\}$, $i$ is the index of the individual in the population, $j'$ represents the treatment site, $\alpha_{j'}$ is the fixed slope of dose for treatment $j'$ to mutated individual of $\dot{\chi}_{i,t}$ at generation $t$, and $\beta_{j'}$ is random slope of dose for parent vector of $\chi_{i,t}$ as the $i^{th}$ individual at generation $t$, and finally $\mathcal{F}_c \in [0,1]$ is the crossover factor which is set by the user. Accordingly, the crossover operation is defined as follows:

$$\ddot{\chi}_{i,t} = \alpha_j \dot{\chi}_{i,t} + \beta_j \chi_{i,t} + e_{j,i,t} \quad (11)$$

$\dot{\chi}_{i,t}$ and $\chi_{i,t}$ are the mutated and the parent solution vectors, respectively. $e_{j,i,t}$ is error term following normal distribution with mean of 0.

*iv) Evaluation and Selection*

The offspring produced by the crossover and mutation operations is evaluated using defined objective function $Q_{total}$, as shown in (12). The best fitted solutions produced by evolution operators are selected and transferred to the next generation ($t + 1$) and the rest will be eliminated.

$$\begin{aligned} \dot{\chi}_{i,t+1} &= \begin{cases} \dot{\chi}_{i,t} & Q_{total}(\chi_{i,t}) \leq Q_{total}(\dot{\chi}_{i,t}) \\ \chi_{i,t} & Q_{total}(\chi_{i,t}) > Q_{total}(\dot{\chi}_{i,t}) \end{cases} \\ \ddot{\chi}_{i,t+1} &= \begin{cases} \ddot{\chi}_{i,t} & Q_{total}(\chi_{i,t}) \leq Q_{total}(\ddot{\chi}_{i,t}) \\ \chi_{i,t} & Q_{total}(\chi_{i,t}) > Q_{total}(\ddot{\chi}_{i,t}) \end{cases} \end{aligned} \quad (12)$$

This results in an iterative improvement of the solutions. The following pseudocode shows how the DE is applied for multi-UAVs mission planning problem.

## 4. Simulation Results and Discussion

The de-centralized mission planner in this research, uses a priori information of sensors distributions (described in Section 2), maximum operation time and battery capacity of each UAV to compute the most appropriate order of tasks (from start point toward the end destination). The planner is able to identify and decide the sensors that should be abandoned, while considering the time threshold constraint (as explained in Algorithm (1)) and sensors' priority index. This section aims at validating the performance of DE-based multi UAVs collaborative mission planning against non-collaborative method in the same environmental setup. To this end, a benchmark of multi-UAVs collaborative and non-collaborative mission planning method is conducted as follows:

i) *Non-Collaborative* mode: in this mode, the UAVs do not communicate to each other and each vehicle individually plans its mission regardless of the others' mission and only concentrate on the assigned operation zone (allocated sensor set); while the catching order of the sensors take place in a way to minimize the route length and mission time. The main drawback of non-collaborative operation is that if any vehicle completes its mission with considerable remaining time, it cannot use this time to help other vehicles in completing their tasks.

ii) *Collaborative* mode: in this mode, the UAVs can communicate each other and exchange information of their pose and task completed as well as managing the endurance time to visit more sensors. In this case, if any vehicle gets discarded from the team (abort its mission for any reason) or runs out of battery, the closest vehicle(s) with the most similar configurations undertake(s) the incomplete mission or the mission is divided between the operational vehicles and the planner module re-plans a new mission scenario for them. This leverages mission coverage range, and system resilience

| **Algorithm (1) – Pseudocode of DE algorithm** |
|---|
| *Inputs:* |
|    Population size ($i_{max}$), Population Index ($i$), |
|    Maximum iteration ($t_{max}$), Number of UAVs ($q$), |
|    Maximum available time ($\mathcal{T}_{\mathcal{U}_j}^{max}$) for vehicle $\mathcal{U}_j$, |
|    number of distributed sensors ($\mathcal{S}_{xy}$) in the area ($n$) |
|    Number of the met sensors in a route ($n'$) |
|    the network topology adjacency matrix ($\mathcal{Cp}_{xy}$) |
| 1.   //*Initialization* |
| 2.   Initialize solution vectors $\chi_{i,t}$ randomly with uniform probability restricted to priority values and $\mathcal{Cp}_{xy}$: <br> $i = 1, \dots, i_{max}$ <br> $t = 1, \dots, t_{max}$ <br> $\chi_{i,t} = (\chi_{1,1} \quad \chi_{2,1} \quad \cdots \quad \chi_{i_{max},1})$ ; <br> Set the mutation coefficient <br> Set the crossover coefficient |
| 3.   **For** $i = 1$ to $i_{max}$ |
| 4.     $ind(i) = \text{cluster}(\{\mathcal{S}_{x,y}^i \in \mathcal{Cp}_{xy}\}, q)$ |
| 5.     **For** $j = 1$ to $q$ |
| 6.       $\text{get}(ind(i,j))$ |
| 7.       $\mathcal{Q}_{total}(i,j) = \mathcal{Q}_{total}(ind(i,j))$ |
| 8.       $\chi_{i,1}(\mathcal{U}_j): \text{return}(ind(i,j))$ |
| 9.     **end For** |
| 10.  **end For** |
| 11.  // *DE main loop* |
| 12.  **For** $i = 1$ to $i_{max}$ |
| 13.   **For** $t = 1$ to $t_{max}$ |
| 14.     **For** $j = 1$ to $q$ |
| 15.       //*Mutation* |
| 16.       $\dot{\chi}_{i,t}(\mathcal{U}_j) = \chi_{r3,t} + \mathcal{F}_\mathcal{M}(\chi_{r1,t} - \chi_{r2,t})$ <br>       $\{r1 \neq r2 \neq r3\} \in \{i = 1, \dots i_{max}\}$ |
| 17.       $\mathcal{Q}_{total}(i,j,t) = \mathcal{Q}_{total}(\dot{\chi}_{i,t}(\mathcal{U}_j))$ |
| 18.       **if** $\mathcal{Q}_{total}(\dot{\chi}_{i,t}(\mathcal{U}_j)) \geq \mathcal{Q}_{total}(\chi_{i,t}(\mathcal{U}_j))$ |
| 19.         $\dot{\chi}_{i,t+1}(\mathcal{U}_j) = \dot{\chi}_{i,t}(\mathcal{U}_j)$ |
| 20.       **else if** $\mathcal{Q}_{total}(\dot{\chi}_{i,t}(\mathcal{U}_j)) < \mathcal{Q}_{total}(\chi_{i,t}(\mathcal{U}_j))$ |
| 21.         $\dot{\chi}_{i,t+1}(\mathcal{U}_j) = \chi_{i,t}(\mathcal{U}_j)$ |
| 22.       **end if** |
| 23.       //*Crossover* |
| 24.       $\forall j' \in 1, \dots, m;\ \ddot{\chi}_{i,t}(\mathcal{U}_j) = \begin{cases} \alpha_{j'}\dot{\chi}_{i,t}(\mathcal{U}_j) & \alpha_{j'} \leq \mathcal{F}_c \vee j' \\ \beta_{j'}\chi_{i,t}(\mathcal{U}_j) & \beta_{j'} \leq \mathcal{F}_c \wedge j' \end{cases}$ <br> $m \in [1, i_{max}]$ |
| 25.       $\ddot{\chi}_{i,t}(\mathcal{U}_j) = \alpha_j \dot{\chi}_{i,t}(\mathcal{U}_j) + \beta_j \chi_{i,t}(\mathcal{U}_j) + e_{j,i,t}$ |
| 26.       $\mathcal{Q}_{total}(i,j,t) = \mathcal{Q}_{total}(\ddot{\chi}_{i,t}(\mathcal{U}_j))$ |
| 27.       **if** $\mathcal{Q}_{total}(\ddot{\chi}_{i,t}(\mathcal{U}_j)) \geq \mathcal{Q}_{total}(\chi_{i,t}(\mathcal{U}_j))$ |
| 28.         $\ddot{\chi}_{i,t+1}(\mathcal{U}_j) = \ddot{\chi}_{i,t}(\mathcal{U}_j)$ |
| 29.       **else if** $\mathcal{Q}_{total}(\ddot{\chi}_{i,t}(\mathcal{U}_j)) < \mathcal{Q}_{total}(\chi_{i,t}(\mathcal{U}_j))$ |
| 30.         $\ddot{\chi}_{i,t+1}(\mathcal{U}_j) = \chi_{i,t}(\mathcal{U}_j)$ |
| 31.       **end if** |
| 32.       //*Collaboration* |
| 33.       **while** $\mathcal{T}_{\mathcal{U}_j} < \mathcal{T}_{\mathcal{U}_j}^{max}$ |
| 34.         Get the label of the abandoned sensors $l = \{\mathcal{S}_{xy}^{i'}\}$ |
| 35.         Get the position of the abandoned sensors |
| 36.         Get the absolute velocity of the idle UAV $|\mathcal{V}_{g_{\mathcal{U}_j}}|$ |
| 37.         $n' = \text{size}(l)$ |
| 38.         $l' = \text{index}\{l\}$ |
| 39.         **For** $k = 1$ to $n'$ |
| 40.           get $\mathcal{S}_{xy}^k$ |
| 41.           $\mathcal{D}'_{\mathcal{S}_{xy}^k}(\mathcal{U}_j) = \text{dist}(\mathcal{S}_{xy}^k, \mathcal{U}_{j,(xy)})$ |
| 42.           Sort ascending $\mathcal{D}_{\mathcal{S}_{xy}^k}(\mathcal{U}_j)$ |
| 43.         **end For** |
| 44.         $\mathcal{S}_{xy}^{l'}(best) = \min(\mathcal{D}_{\mathcal{S}^{l'}_{xy}})$ |
| 45.         $\mathcal{T}_{\mathcal{U}_j}(\mathcal{S}_{xy}^{l'}(best)) = \sum \mathcal{T}_{\mathcal{U}_j} + \frac{\mathcal{D}'(\mathcal{U}_j)}{\mathcal{V}_{g_{\mathcal{U}_j}}(\rho^{i'})}$ |
| 46.         Add $\mathcal{S}_{xy}^{l'}(best)$ to $\mathcal{U}_j$ route |
| 47.       **end while** |
| 48.     **end For** ($j$) |
| 49.   **end For** ($t$) |
| 50.  **end For** ($i$) |
| *Output:* |
|    The shortest route for each UAV $\mathcal{U}_i$ with best order of prioritized sensors |

To investigate the performance of the proposed mission planner, series of experiments are conducted. To this end, four homogeneous UAVs are deployed to a bush area, and all vehicles are configured with the same setting. The battery lifetime threshold for each vehicle is equally set to $3.6 \times 10^3 (sec)$ for a continuous operation. The dimension of operation field is considered in a scale of $(3.4_x \times 3.4_y)$ kilometres and 128 sensors are non-uniformly distributed in the operation field, as depicted in Fig.5. UAVs take advantage of partially constructed map to have a perception of the operating field in a priori, departure from different Start positions, and after the completion of the mission they should arrive at the same End point (a priori known coordinate). In this study, all computations were performed on a desktop PC with an Intel i7 3.20 GHz quad-core processor in MATLAB®2019a.

The sensor nodes are divided into ($k=q$) different groups according to the sensors collection priority and distribution, where the highest priority sensors are grouped in smaller clusters to ensure the assigned UAV will have sufficient time to serve them. In Fig.5 (b), colors of sensor node represent different SN-groups.

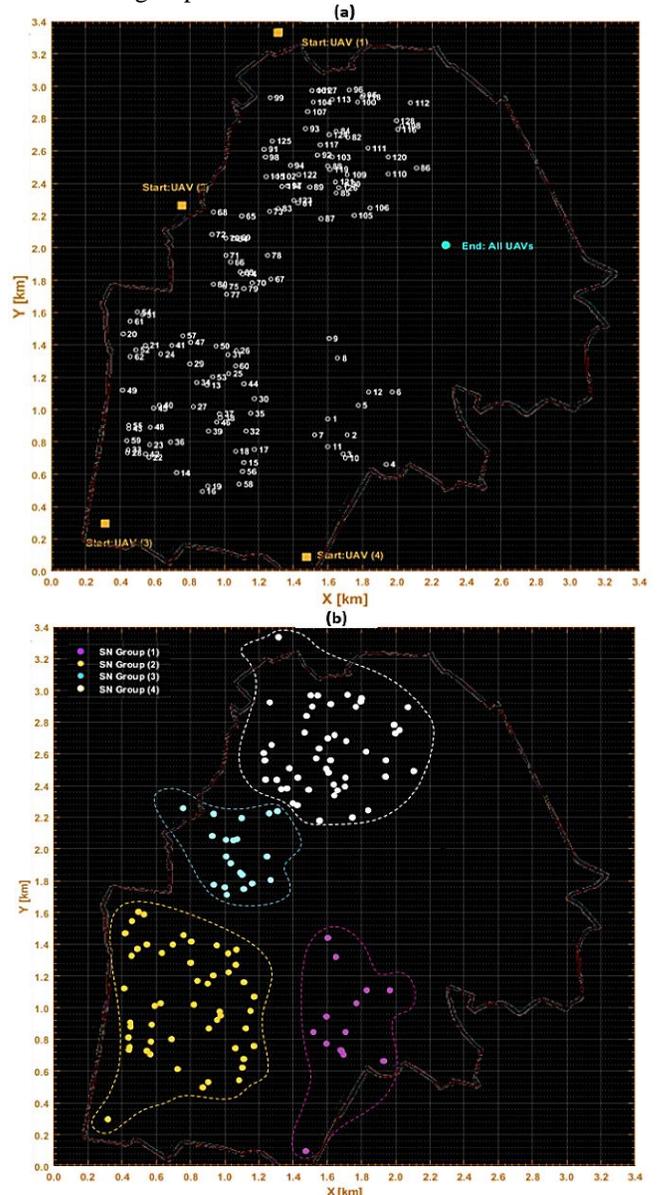

Fig. 5. (a) Distribution of 128 sensors, 4 departure (start) points and an identical landing (end) location (b) SN-groups produced by precedence-based k-means.

Figures 6 shows the performance of the mission planner for meeting the maximum number of sensors in both collaborative and non-collaborative modes. In this figure, the sensors are labeled from 1 to 128 and the mission planner divides them between four UAVs. As demonstrated in Fig.6, all distributed sensors are grouped-up by the mission planner in accordance with number of vehicles and sensors location (four groups for four vehicles). As shown in the figure, the SN groups of (2) and (4) encompass a larger area with a higher number of sensors, while the other two SN clusters encapsulate fewer number of sensors. This difference distinguishes how the collaborative mode enables vehicles to undertake others' tasks in their spare time. The proposed mission is conducted in a non-collaborative mode while there is no link of communication between UAVs whereas in the collaborative mode the vehicles can communicate each other and share/exchange the information of mission. These include transmitting the mission updates, current location coordinates, and the visited sensors to enhance the coverage and to avoid redoing the task.

As shown in Fig.6 (a), in the non-collaborative mode, all the vehicles only concentrate on their assigned SN group without being aware of other vehicles' operation details, and accordingly the operation zone for each UAV is smaller compared to the collaborative mode in Fig.6 (b). As a result, the $1^{st}$ and $3^{rd}$ UAVs can meet all the allocated sensors as the corresponding SN groups encapsulate a fewer number of sensors. However, in the $2^{nd}$ and $4^{th}$ SN groups, a considerable number of sensors are abandoned by UAV-2 and UAV-4 due to the shortage of time (as all the vehicles are selected from the same class with the same battery capacity). Although the battery life-time threshold is not violated by the UAVs when they arrive at the rendezvous point, the $1^{st}$ and $3^{rd}$ UAVs did not use all their capacity at the mission completion, while they could effectively use their residual time for assisting the other vehicles after completing their own mission. In contrast, considering the given

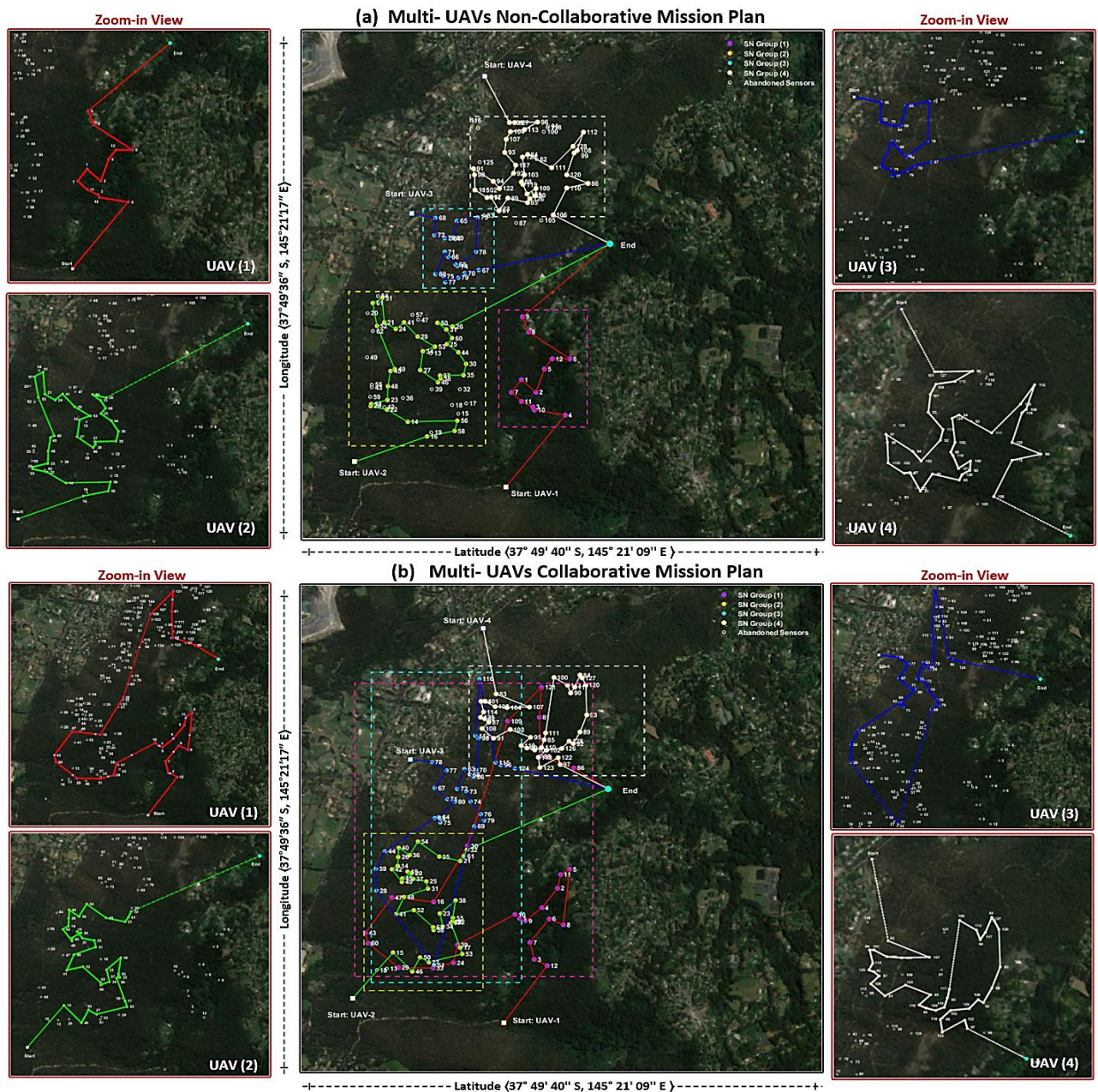

Fig.6 (a) The route taken by multiple UAVs in non-collaborative mode; (b) The route taken by multiple UAVs in collaborative mode.

data acquisition pattern in the collaborative mission planning (see Fig.6 (b)), the UAVs operation zones are extended to a larger area, where the 1st and 3rd UAVs successfully meet all the existing sensors in SN-groups (1) and (3) and devote their remained time to cooperate with the 2nd and 4th vehicles for accomplishing the abandoned sensors in the neighborhood.

Figure 7 is more clear demonstration of Fig.6 (without background) highlighting the mission undertaken in non-collaborative and collaborative modes. As shown in Fig.7 (a) and (b), there is no significant difference in terms of the data acquisition pattern, quantity of met sensors and the size of operation zone for 2nd and 4th UAVs in the collaborative and non-collaborative methods. This is due to the fact that the corresponding SN-groups (2 and 4) already comprise many sensors and thus the vehicles' battery-life cannot be sufficient to meet all of them. As a result, the 2nd and 4th UAVs will not show extra time to overtake any sensor from nearby SN-groups. In contrast, the 1st and 3rd UAVs that assigned to smeller SN-groups (1 and 3), devote their extra time to help the 2nd and 4th UAVs. To be precise, it is shown in Fig.7 (b) that the 1st UAV flies toward SN-group (2), as the closest neighbor, after accomplishing its operation zone of SN-group (2). Then the UAV-1 skips the 3rd SN-group as all the area is already completed by the 3rd UAV, and picks some of the sensors from SN-group (4) and moves toward the rendezvous point before running out of battery. As a result, the operation zone of the 1st UAV in Fig.7 (b), is greatly extended in the collaborative mode compared to the other scenario in Fig.7 (a). Similarly, the 3rd UAV first completes its own area, then extend its operation zone by flying toward SN-group (2), as the closest neighbor, and SN-group (4), as the area on the way of rendezvous point. This is a good indication of how collaboration impacts taking the best use of available resources and coverage of an extended area without extra resource allocation for the same mission.

Figures 8 and 9 demonstrate the evaluation of the mission planner based on the performance indices and penalty functions such as augmented mission cost (indicated in (4)), individual mission cost per UVA, time difference between battery life-time and the operation time of each UAV(given by (13)), and mission time violation. It should be noted that since the evaluation criterion for deciding the best solution is the total cost of the four UAVs operation, in some iterations one vehicle may experience a slight increase in cost value comparing with the previous iterations; but the summation of the costs keeps reducing iteratively.

$$\forall \mathcal{U}_j \in \mathcal{U}_1, \ldots, \mathcal{U}_q, \quad \mathcal{T}_{diff} = \mathcal{T}_{\mathcal{U}_j}^{\max} - \mathcal{T}_{\mathcal{U}_j} \quad (13)$$

The total cost trend in Fig. 9(a) shows that the DE-based mission planner effectively converges the total collaborative mission cost of $1.415 \times 10^3$, which is remarkably less than the non-collaborative mission cost of $2.246 \times 10^3$ in Fig. 8(a). In both operations, the planner tends to maximize the met sensors and minimize the route length, while taking time constraint into account. However, the non-collaborative operation does not consider the time difference between battery life-time and the operation time of each UAV, which results in a greater cost as UAVs 1 and 3 finish their mission with a considerable residual battery. In contrast, in collaborative mode, the UAVs communicate mission updates with each other and overtake the others mission in their remaining time and this leads to a further cost minimization.

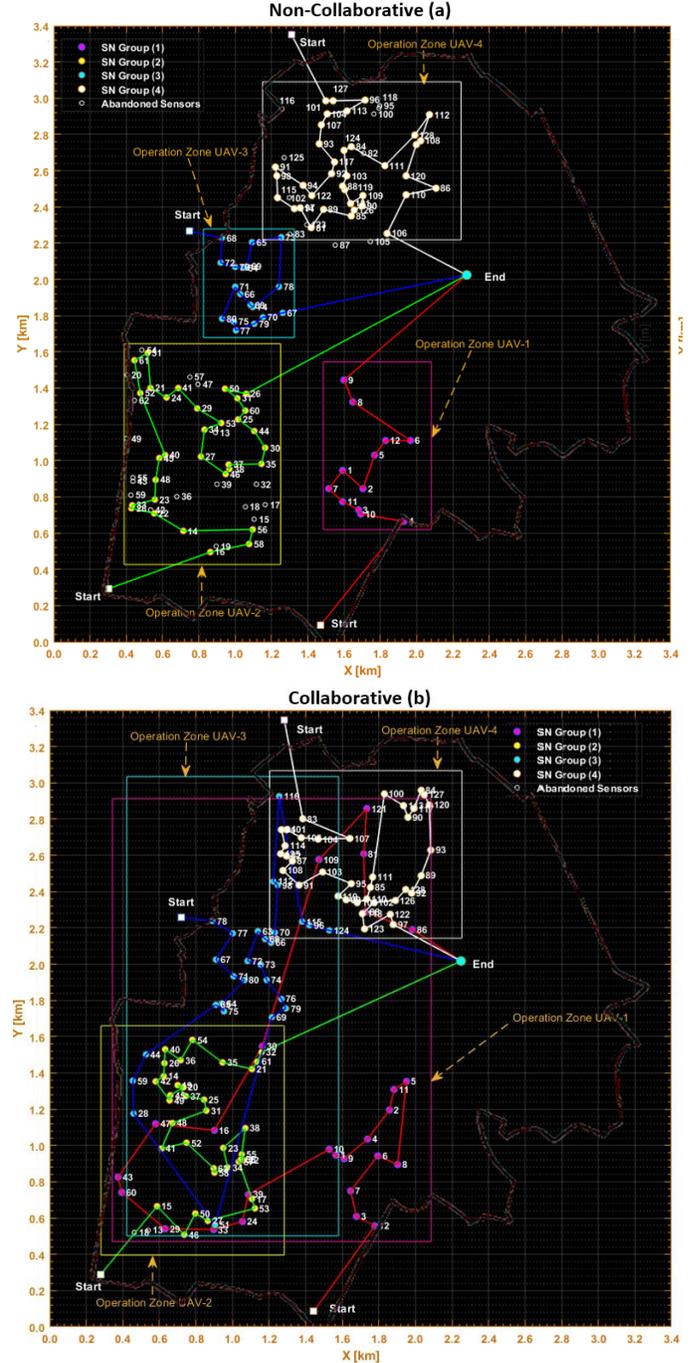

Fig.7 (a) performance demonstration of the mission planner in the non-collaborative and (b) collaborative modes.

The time constraint enforces the vehicles to abandon some of the sensors in the crowded SN-groups. To be more precise on individual UAVs operation cost, it is notable from both Fig. 8(b) and Fig.9 (b) that the best performance in minimizing the cost belongs to the 2nd and 4th UAVs while for other two UAVs (1 and 3), the DE-based mission planner is converged to the best solution in beginning iterations. This is due to the fact that these two UAVs do not have any time violation and complete their assigned SN-groups (which are smaller than SN-groups of 2 and 4) just with a portion of their battery lifetime.

Considering the violation diagram in Fig.8(c) and Fig.9(c), the DE-based mission planner manages the time violation for both collaborative and non-collaborative modes by leaving the least priority sensors (to fit the operation time to the maximum available time). By continuing this process, the DE-based planner eliminates the violation iteratively. As illustrated in Fig.8 (c) and Fig.9(c), the 1st and 3rd UAVs do not violate the time constraint over the entire iterations while the 2nd and 4th UAVs have experienced a high level of violation at starting iterations; however, the algorithm gradually reduces the violation value and converges the zero after 110 iterations.

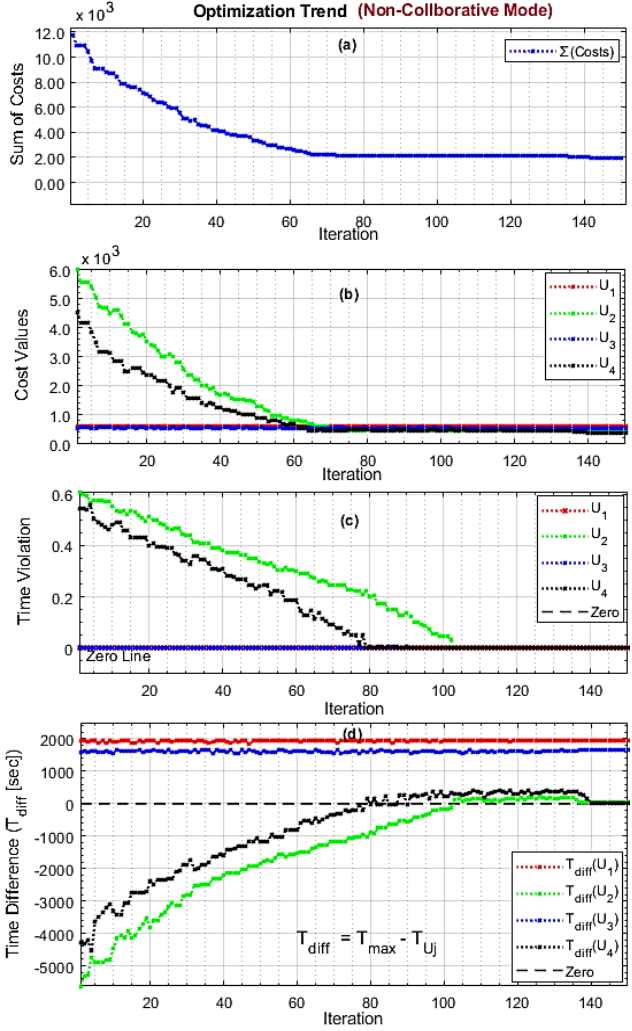

Fig.8. Evaluation of the DE-based mission planner in satisfying the mission performance criteria for multi UAVs non-collaborative data acquisition; (a) total mission cost; (b) individual UAV's cost variation (c) UAVs' violation of battery life-time; and (d) mission time difference ($\mathcal{T}_{diff}$) for individual UAV's.

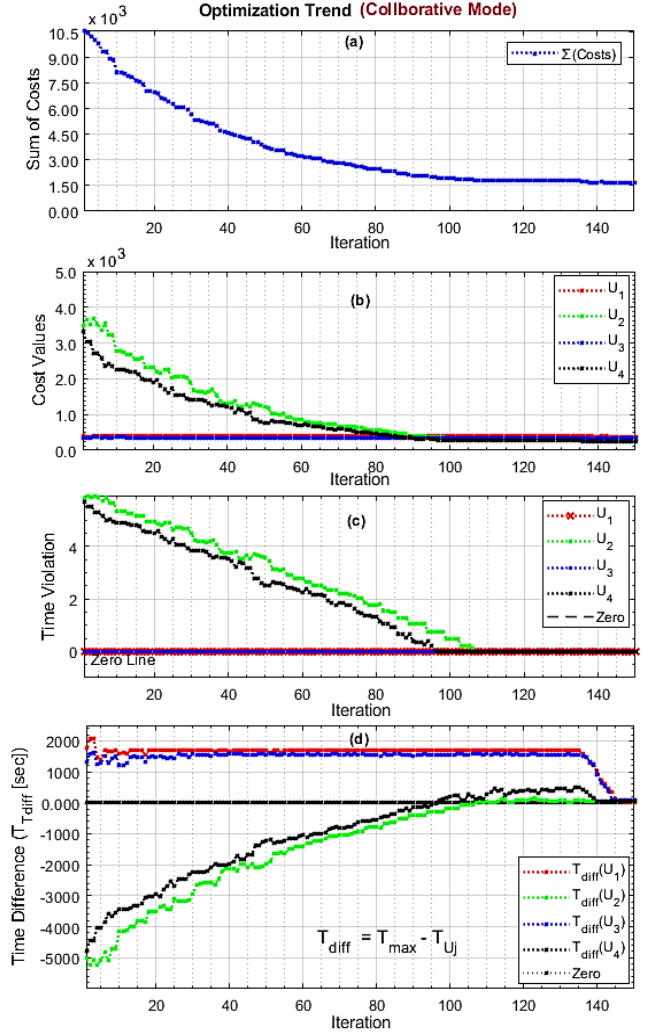

Fig.9. Evaluation of the DE-based mission planner in satisfying the mission performance criteria for multi UAVs collaborative data acquisition; (a) total mission cost; (b) individual UAV's cost variation (c) UAVs' violation of battery life-time; and (d) mission time difference ($\mathcal{T}_{diff}$) for individual UAV's.

Referring to Fig.8 (d), after all iterations are completed, the time-difference ($\mathcal{T}_{diff}$) for 1st and 3rd UAVs remains with a large positive value (approximately +1800 to +2000 *sec*), and they cannot entirely eliminate the time difference due to the lack of collaboration mechanism. On the other hand, the 2nd and 4th UAVs start with a great time-difference violation (about −4000 to −5000 *sec*) that means these two UAVs experience the shortage of time for accomplishing their missions. In this situation, the DE-based mission planner suppresses the $\mathcal{T}_{diff}$ iteratively by eliminating least priority sensors from mission plan. Consequently, the less number of sensors are visited due to the lack of collaboration.

Considering Fig.9 (d), the collaboration mechanism facilitates the algorithm to assign the abandoned sensors to the idle vehicles with spare time and enforces both negative and positive $\mathcal{T}_{diff}$ to approach zero. This means the 1st and 3rd UAVs in the collaborative mode, devote their positive (residual) time to reduce the load of other vehicles that results in taking the best use of total available time. The time difference in Fig.9 (d) and the violation graphs in Fig.9 (c) thoroughly endorse the performance of the proposed DE-based mission planner in the collaborative mode.

Table 1 provides a detailed numerical comparison of the DE-based mission planner performance for the collaborative and non-collaborative scenarios.

From the numerical point of view, the total time difference for the non-collaborative planner is +3,675 (sec) while the in collaborative operation, the $\mathcal{T}_{diff}$ is +210 (sec) out of total available time 14400 (sec) (4×3.6×10³) for four UAVs. This results in that the collaborative planner collectively visits 126 out of 128 sensors while this number is noticeably decreased to 99 sensors with non-collaborative planner. The reason is that the non-collaborative planner only enables the vehicles to individually optimize the route length, sensors' order and quantity. This enforces the UAVs in larger SN-groups to ignore most of the sensors to meet time restriction, while the UAVs in smaller SN-groups can complete all their tasks in a small portion of their available time and finish their mission without considering the nearby sensors (as no restriction confines the planner to minimize the time-difference).

Table 1. Numerical assessment and comparison of performance indices in non-collaborative and collaborative scenarios

|  | Non-Collaborative | Collaborative |
|---|---|---|
| **Operation Cost** $(Q(\mathcal{U}_j)^{-1})$ | $\mathcal{U}_1 \cong 0.698 \times 10^3$ <br> $\mathcal{U}_2 \cong 0.452 \times 10^3$ <br> $\mathcal{U}_3 \cong 0.583 \times 10^3$ <br> $\mathcal{U}_4 \cong 0.413 \times 10^3$ <br> $\sum \mathcal{U}_j = 2.14 \times 10^3$ | $\mathcal{U}_1 \cong 0.419 \times 10^3$ <br> $\mathcal{U}_2 \cong 0.311 \times 10^3$ <br> $\mathcal{U}_3 \cong 0.387 \times 10^3$ <br> $\mathcal{U}_4 \cong 0.298 \times 10^3$ <br> $\sum \mathcal{U}_j = 1.41 \times 10^3$ |
| **Operation Duration** $(\mathcal{T}_{\mathcal{U}_j})$ | $\mathcal{U}_1 = 1{,}665.00$ (sec) <br> $\mathcal{U}_2 = 3{,}538.00$ (sec) <br> $\mathcal{U}_3 = 1{,}945.00$ (sec) <br> $\mathcal{U}_4 = 3{,}577.00$ (sec) <br> $\sum \mathcal{U}_j = 10{,}725$ (sec) | $\mathcal{U}_1 = 3{,}581.00$ (sec) <br> $\mathcal{U}_2 = 3{,}568.00$ (sec) <br> $\mathcal{U}_3 = 3{,}504.00$ (sec) <br> $\mathcal{U}_4 = 3{,}537.00$ (sec) <br> $\sum \mathcal{U}_j = 14{,}292.0$ (sec) |
| **Time Difference** $(\mathcal{T}_{diff} = \mathcal{T}_{\mathcal{U}_j}^{max} - \mathcal{T}_{\mathcal{U}_j})$ | $\mathcal{U}_1 = +1935.0$ (sec) <br> $\mathcal{U}_2 = +62.000$ (sec) <br> $\mathcal{U}_3 = +1655.0$ (sec) <br> $\mathcal{U}_4 = +23.000$ (sec) <br> $\sum \mathcal{U}_j = +3{,}675$ (sec) | $\mathcal{U}_1 = +19.00$ (sec) <br> $\mathcal{U}_2 = +32.00$ (sec) <br> $\mathcal{U}_3 = +96.00$ (sec) <br> $\mathcal{U}_4 = +63.00$ (sec) <br> $\sum \mathcal{U}_j = +210.0$ (sec) |
| **Visited Sensors** $(\sum S^i \in C_p)$ | $\mathcal{U}_1 = 12$ <br> $\mathcal{U}_2 = 32$ <br> $\mathcal{U}_3 = 18$ <br> $\mathcal{U}_4 = 37$ <br> $\sum \mathcal{U}_j = 99/128$ | $\mathcal{U}_1 = 26$ <br> $\mathcal{U}_2 = 35$ <br> $\mathcal{U}_3 = 28$ <br> $\mathcal{U}_4 = 37$ <br> $\sum \mathcal{U}_j = 126/128$ |
| **Travelled Distance** $(\mathcal{D}_{\mathcal{U}_j})$ | $\mathcal{U}_1 = 825.00$ (m) <br> $\mathcal{U}_2 = 1{,}498.0$ (m) <br> $\mathcal{U}_3 = 745.00$ (m) <br> $\mathcal{U}_4 = 1{,}237.0$ (m) <br> $\sum \mathcal{U}_j = 4{,}305.0$ (m) | $\mathcal{U}_1 = 1{,}901.00$ (m) <br> $\mathcal{U}_2 = 1{,}348.00$ (m) <br> $\mathcal{U}_3 = 1{,}704.00$ (m) <br> $\mathcal{U}_4 = 1{,}197.00$ (m) <br> $\sum \mathcal{U}_j = 6{,}150.0$ (m) |
| **Time Violation** $(\mathcal{T}_{\mathcal{U}_j} - \mathcal{T}_{\mathcal{U}_j}^{max} \geq 0)$ | $\mathcal{U}_1 = 0.0000$ <br> $\mathcal{U}_2 = 0.0000$ <br> $\mathcal{U}_3 = 0.0000$ <br> $\mathcal{U}_4 = 0.0000$ <br> $\sum \mathcal{U}_j = 0.0000$ | $\mathcal{U}_1 = 0.0000$ <br> $\mathcal{U}_2 = 0.0000$ <br> $\mathcal{U}_3 = 0.0000$ <br> $\mathcal{U}_4 = 0.0000$ <br> $\sum \mathcal{U}_j = 0.0000$ |

As indicated in Table 1, in the non-collaboration, UAVs ($\mathcal{U}_1$) and ($\mathcal{U}_3$) terminate the mission with a large positive time difference of +1935.0 (sec) and +1655.0 (sec), respectively. It should be noted that the cost value has a direct relation to the violation and number of the visited sensors and therefore, the cost value for the non-collaborative mission ($\sum \mathcal{U}_j = 2.14 \times 10^3$) is considerably higher than the collaborative one ($\sum \mathcal{U}_j = 1.41 \times 10^3$). Due to the smaller size of SN-group (1) that comprise only 12 sensors, the $\mathcal{U}_1$ visits all the 12 sensors in its mission with a largely positive $\mathcal{T}_{diff}$, where this time can be used to take the negative load of other UAVs. As a result, the non-collaborative planner experiences a higher cost compared to the collaborative planner. Finally, there is no violation in both scenarios that means the fleet of UAVs completed their missions successfully before running out of battery. This confirms the feasibility of the produced routes.

## 5. Conclusions

This study proposed an effective UAV-based data collection strategy that can be applied to monitor and collect the data of a large distributed sensor network. This was established upon developing a collaborative mission planner system enabling a team of UAV to efficiently conduct and complete the sensors' data collection mission considering several payload and physical constraints such as limited battery capacity of UAVs for long term missions. The proposed mission planner system utilized DE optimization engine to properly suggest the optimal routes and orders for maximization of the number of visited sensor nodes at the end of mission.

The performance of multi-UAVs collaborative mission planning was investigated within extensive simulation studies and compared to the non-collaborative mission planning. The simulation results declare that by using collaborative mission planner, the operation cost of the mission can be reduced up to 35.12% when it is compared to the non-collaborative planning. Moreover, data collection of 98.43% whole network becomes possible by utilizing the proposed collaborative mission planner while this number is decreased to 77.34% for the non-collaborative operation. This is a good indication of how the DE-based collaborative planner optimally uses all UAVs battery capacity. The result of comparative analysis confirmed the effectiveness of the collaborative planner in mission time management and maximum completion of the available tasks (i.e., sensor vesting and data collection) in the mission. Future extension of this study includes field trials and evaluation of the planner in practice.

## Conflicts of Interest

The authors declare that this article does not contain any conflict of interest.

IEEE Transactions on Cybernetics, vol. (99), pp.1-14, 2018. DOI: 10.1109/TCYB.2018.2837134.
[16] V. Roberge, M. Tarbouchi, G. Labonte. Comparison of Parallel Genetic Algorithm and Particle Swarm Optimization for Real-Time UAV Path Planning. IEEE Transactions on Industrial Informatics, vol.9(1), pp.132–141, 2013. doi: 10.1109/TII.2012.2198665
[17] S. MahmoudZadeh, D. Powers, R.B. Zadeh. Autonomy and Unmanned Vehicles "Augmented Reactive Mission–Motion Planning Architecture for Autonomous Vehicles. Springer Nature, Cognitive Science and Technology, 2019. ISBN 978-981-13-2245-7. DOI: 10.1007/978-981-13-2245-7
[18] V.G.Tonge. Travelling Salesman Problem using Differential Evolutionary Algorithm International Conference on Innovation & Research in Engineering, Science & Technology (ICIREST-19), vol.63, pp. 63-67, 2019.
[19] K. Price, R. Storn. Differential evolution-A simple evolution strategy for fast optimization. Dr. Dobb's Journal, vol.22 (4), pp.18-24, 1997.## Declarations

The authors declare that the current research is new in contribution and the content is not published elsewhere.

*Funding information*: Not applicable.

*Conflicts of interest*: the authors declare that this article does not contain any conflict of interest.

*Ethics approval*: Not applicable.

*Availability of data and material*: Not applicable

*Code availability*: the code can be available on demand. In this study, all computations were performed on a desktop PC with an Intel i7 3.20 GHz quad-core processor in MATLAB®2019a.

*Consent to participate:* The authors would like to submit this manuscript for the possible evaluation in journal of Neural Computing and Applications, Special Issue on CI-based Control and Estimation in Mechatronic Systems.

*Consent for publication:* Its publication has been approved by all co-authors, as well as by the responsible authorities at the institute where the work has been carried out.